\pdfoutput=1

\documentclass[11pt]{article}

\usepackage[]{acl}

\usepackage{times}
\usepackage{latexsym}

\usepackage[T1]{fontenc}

\usepackage[utf8]{inputenc}

\usepackage{microtype}
\usepackage{graphicx}
\usepackage{bbm}
\usepackage{amssymb}
\usepackage{amstext}
\usepackage{booktabs}
\usepackage{multirow}
\usepackage{tabularx}
\usepackage{float}
\usepackage{caption}
\usepackage{amsmath}
\usepackage{url}
\usepackage{marvosym}

\title{Integrating Physician Diagnostic Logic into Large Language Models: Preference Learning from Process Feedback}

\author{
Chengfeng Dou\textsuperscript{1,2}, 
Ying Zhang\textsuperscript{3},
Zhi Jin\textsuperscript{1,2}\Letter, 
Wenpin Jiao\textsuperscript{1,2}\Letter, 
Haiyan Zhao\textsuperscript{1,2}, \\
\textbf{Yongqiang Zhao\textsuperscript{1,2}}, 
\textbf{Zhengwei Tao\textsuperscript{1,2}}\\
\textsuperscript{1} School of Computer Science, Peking University;\\ 
\textsuperscript{2} Key Laboratory of High Confidence Software Technologies(PKU), MOE, China \\
\textsuperscript{3} Beijing Key Lab of Traffic Data Analysis and Mining, Beijing Jiaotong University, Beijing, China\\
\texttt{\{chengfengdou,zhijin,jwp,zhhy.sei\}@pku.edu.cn}\\
~\texttt{\{tttzw,yongqiangzhao\}@stu.pku.edu.cn} \texttt{\{19112043\}@bjtu.edu.cn}
}

\begin{document}
\maketitle
\begin{abstract}
The utilization of large language models for medical dialogue generation has attracted considerable attention due to its potential to enhance response richness and coherence. While previous studies have made strides in optimizing model performance, there is a pressing need to bolster the model's capacity for diagnostic logic to ensure patient safety. In response to this need, we propose an approach termed preference learning from process feedback (PLPF), which involves integrating the doctor's diagnostic logic into LLMs.
PLPF encompasses three key components: rule modeling, preference data generation, and preference alignment. These components collectively serve to train the model to adhere to the diagnostic process. Our experimental results, utilizing Standardized Patient Testing, demonstrate that PLPF enhances the diagnostic accuracy of the baseline model in medical conversations by 17.6\%, surpassing the performance of traditional approaches. Moreover, PLPF exhibits effectiveness in both multi-round and single-round dialogue tasks, thereby highlighting its potential in improving medical dialogue generation.
Our dataset is available at \url{https://github.com/Chengfeng-Dou/SpTesting}
\end{abstract}
\section{Introduction}

The use of large language models~(LLMs)~\cite{zhao2023survey} has recently exploded in the field of medical dialogue generation.
However, training robust medical dialogue models is crucially based on high-quality training data~\cite{he2023survey}. 
As a result, considerable efforts have been made to generate extensive training data sets to fine-tune these models. Furthermore, certain studies have made notable progress, such as the application of reinforcement learning from human feedback~(RHLF) to guide models in generating user-friendly responses~\cite{chen2023huatuogpt, bao2023disc}.

Despite the significant advancements in prior research on RLHF, current open-source medical LLMs remain inadequately user-directed. 
They frequently oscillate between two problematic behaviors: 1) precipitously rendering a diagnosis without adequate patient data collection, and 2) becoming ensnared in an endless loop of data gathering without progressing to a definitive diagnosis. 
This issue appears to originate from the inherent discordance among the preference data provided by annotators. 
In practical settings, physicians' diagnostic approaches typically bifurcate into: 1) an initial comprehensive collection of patient information followed by a diagnosis, and 2) an early presentation of a conjectural diagnosis, which is then substantiated through progressive data accumulation. 
It is the simultaneous learning from these two divergent preferences that likely precipitates the disarray in the model's conversational strategy.

\begin{figure*}[t]
    \centering
    \includegraphics[width=0.9\textwidth]{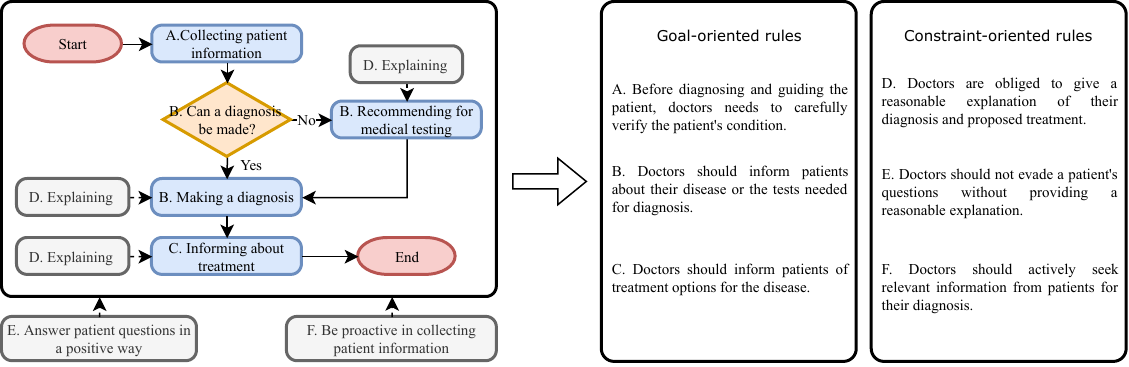}
    \caption{
    Medical diagnosis flowchart~(left) and its corresponding rules~(right). In the flowchart, we use blue boxes for activities, orange diamonds for judgment conditions, and gray boxes for additional constraints. 
    We use the letters A-F to indicate the correspondence between the rules and the elements in the flowchart.
    }
    \label{fig:flow}
\end{figure*}

We believe that standardising the medical dialogue process is the key to solving the above problems. Our proposed approach, known as preference learning from process feedback (PLPF), focuses on ensuring the rationality of the conversation flow, which sets it apart from traditional methods. The core idea is to represent the doctor's diagnostic logic using a flowchart and employ preference learning to train the model to avoid generating responses that deviate from the established process. 

In particular, we have developed a flowchart based on the doctor's expertise, as depicted in Fig.~\ref{fig:flow}. This flowchart effectively outlines the objectives the physician must achieve and the constraints that must be followed during the diagnostic process, while also illustrating the dependencies between these objectives. To utilize the flowchart for guiding model training, we have established explicit rules for each activity, decision, and constraint outlined in the flowchart. The state of a dialogue in the flowchart can be determined by evaluating whether the dialogue conforms to these rules. 

Based on these established rules, our approach consists of three phases: rule modeling, preference data generation, and preference alignment. Initially, we have developed a Rule Evaluation Model (REM) to automatically assess whether a conversation adheres to a specific rule. Building on this, we have devised a method to assign scores to conversations that take into account rule dependencies. These scores are then used for the generation of preference data. We achieve this using an innovative one-shot learning-based approach to predict dialogue trajectories and leveraging REM to appropriately rank these trajectories. Subsequently, we employ the Direct Preference Optimization (DPO) algorithm~\cite{dpo} to train models based on the acquired preference data.

We used standardized patient testing, a widely accepted method in the medical field, to assess our approach. To achieve this, we built the Chinese Standardized Patient Test (CSPT) dataset. Furthermore, we employed a retrieval-augmented generation technique to create a patient simulator for interactive testing with LLM. The results of our experiments indicate that our approach improves the diagnostic precision of the baseline model in medical conversations by 17.6\%. We also tested our approach on three public datasets to assess its performance in both multi-round and single-round conversations. The results show that our approach effectively enhances the model's understanding of physician expressions. In summary, our work provides the following contributions.
\begin{itemize}
    \item We introduce PLPF, a method for multi-round healthcare conversations that allows LLMs to incorporate industry flowchart specifications to improve conversation logic.

    \item We provide a high-quality evaluation dataset for standardized patient testing, offering a novel approach to evaluate the communication skills of medical LLMs.
    
    \item We demonstrate the superiority of PLPF in improving patient diagnostic accuracy through standardized patient testing.
\end{itemize}

\section{Method}
\begin{figure*}
    \centering
    \includegraphics[width=0.9\textwidth]{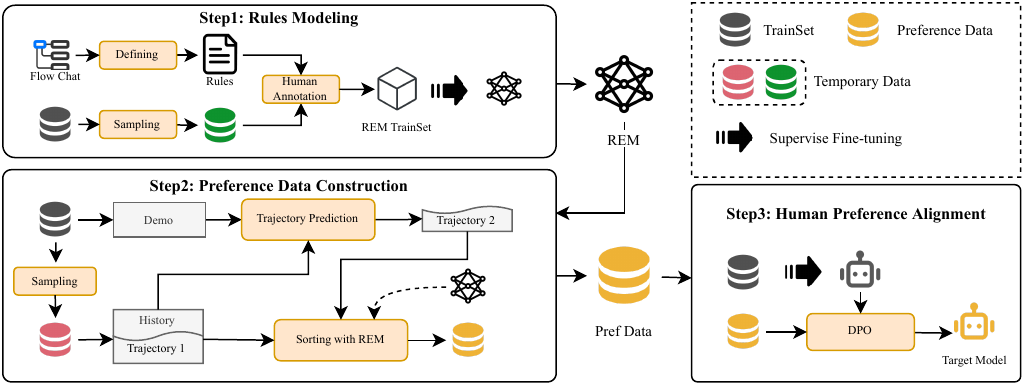}
    \caption{Overview of the training process.
    The training process is divided into three steps, with key activities indicated using orange rounded rectangular boxes. To distinguish the different stages of the data, we labeled them with different colors and provided data descriptions in the upper right corner of the image.
    }
    \label{fig:overview}
\end{figure*}

\subsection{Overview}
The overall training process is depicted in Fig.~\ref{fig:overview}, and consists of three phases: \textit{Rules Modeling}, \textit{Preference Data Construction}, and \textit{Human Preference Alignment}. In the first phase, we establish the corresponding rules using the flowchart, which are then employed by manual annotators to generate a rule evaluation dataset. Afterward, a Rule Evaluation Model~(REM) is developed by training on this dataset. Moving on to the second phase, we initially utilize the REM to filter the training data, ensuring the acquisition of high-quality data. Subsequently, preference datasets are constructed based on the retained data, using ChatGPT and REM. Finally, we employ these preference datasets to train the model, which has been fine-tuned with instruction data, resulting in the final model. Each stage will be described in detail in the following sections.

\subsection{Rules Modeling}
\subsubsection{Rules Definition}
In order to evaluate whether the dialogue follows a specific process, as illustrated in Fig.~\ref{fig:flow}, we need to assess it in two ways: firstly, to determine if the doctor follows the correct sequence to accomplish the goals outlined in the flowchart, and secondly, to verify if the doctor complies with the constraints. To accomplish this, we need to establish specific rules. We have developed a total of six rules for the flowchart, as depicted in Fig.~\ref{fig:flow}. It is important to note that we have categorized these rules into goal-oriented and constraint-oriented rules based on their distinct functions. Goal-oriented rules assess whether the doctor achieves the specified goals, while constraint-oriented rules evaluate how well the goals are achieved. For the sake of clarity in subsequent discussions, we denote the set of these two types of rules as $\mathcal{R}^{g}$ and $\mathcal{R}^{c}$.

\subsubsection{Sampling and Human Annotation}
Once predefined rules are established, it is essential to develop a Rules Evaluation Model (REM) for these rules. The rule evaluation task is structured as a Q\&A format, as shown below.

\vspace{0.2cm}
\noindent \textit{Human: Rule: [Rule]. History: [History].
Did the doctor follow the rule during the conversation?
Assistant: [Comment]. Score: [Score].}
\vspace{0.2cm}

In this template, [Rule], [History], [Comment], and [Score] denote slots to be filled, where [Rule] and [History] are model inputs, while [Comment] and [Score] are model outputs.
For this task, we construct a small training set with hand-crafted annotations to train REM.
In practice, we first collect 400 online medical consultation dialogues and randomly select dialogue segments at different stages.
Subsequently, each data instance is scored and commented on by at least three annotators based on the predefined rules.
To simplify the scoring process, we define the score values as 0, 1, and 2, representing non-compliance, partial compliance, and complete compliance with the rule, respectively. For more information on scoring, please see Appendix~\ref{appendix: scoring}. In the end, we obtained a total of 2,400 samples, which were divided into 1,800 for training and 600 for testing.

During the model training phase, we utilize an auto-regressive training method. 
It is important to note that when calculating the loss function, we only compute the loss value of tokens that appear after the string `Assistant'.

\subsection{Preference Data Construction}
\subsubsection{Basic Idea} 
In this section, we present a general overview of how REM can be used to guide model training. Our goal is to train the model to follow a specific conversational process when interacting with patients. However, it is important to note that real-life conversations may require deviations from this process to address the patient's needs. Therefore, relying solely on the REM score predicted as a reward for reinforcement learning may result in the generation of responses that lack fluency and coherence. To address this issue, we have chosen to employ a contrastive learning-based approach for model training. Specifically, we provide the model with two different candidate responses for the same conversation history, both of which should be fluent and reasonable. We then use REM to guide the model toward learning the response that aligns more closely with the conversational flow, while rejecting the other response. In the following sections, we will provide more details on how candidate responses are generated and how REM is utilized in this process.

\subsubsection{Candidate Responses Generation}
The medical conversation task is complex and consists of multiple stages, making it challenging to achieve all objectives in a single round of conversation. Because there are dependencies between goals, evaluating candidate responses requires integrating information from future rounds of conversations. Therefore, in addition to producing candidate responses, it is necessary to generate dialog trajectories for future rounds of conversations, which can aid in subsequent evaluation. Traditional RLHF-based approaches typically generate candidate responses using a model that has undergone fine-tuning. However, this model has limitations in generating future dialog trajectories as it can only predict one step of future responses. As a result, we incorporate data sampling and trajectory prediction to generate candidate responses.

\paragraph{Data Sampling.}
Sampling from web dialogue data is the most direct approach to acquiring candidate responses. We selected 4,000 samples from the MedDialogue dataset in a random manner. Subsequently, we randomly divide the dialog records of these samples into two sections, enabling us to capture the conversation history, doctor's responses, and future interactions simultaneously.

\paragraph{Trajectory Prediction.} \label{sec:tp}
Trajectory prediction is another way to generate candidate responses. In this research, we use ChatGPT for this purpose. We observed that ChatGPT's training dataset probably includes MedDialogue. The trajectories generated by ChatGPT closely resemble the actual dataset. To distinguish between the two conversation trajectories, we intervene in ChatGPT's prediction process using a one-shot learning approach. We utilize the following instruction templates.

\vspace{0.2cm}
\noindent \textit{You are a dialogue continuation AI, please read the below two dialogues and follow my instructions. \\
Dialogue A: [Arbitrary medical dialogue]. \\
Dialogue B: [The dialogue to be completed]. \\
Please continue Dialogue B while fulfilling the following requirements: \\
1. The doctor's style should match the doctor's style in Dialogue A. \\
2. The patient's style should match the patient's style in Dialogue B.
}
\vspace{0.2cm}

\subsubsection{Sorting with REM}
After acquiring the candidate responses and their respective future conversation trajectories, the REM is utilized to score each candidate response for ranking purposes. To fully utilize the information from future conversation trajectories, the following formula is adopted for computing the score. 
\begin{equation}
s(c \mid h) = v(h, c) + \sum_{i=1}^{n} d^i v(h, c, ..., u_i, a_i)
\end{equation} 
In the provided equation, the variable $s$ represents the score of candidate responses, while $v$ represents the score of the conversation states, which is implicitly embedded in the conversation history and the doctor's responses. The variable $h$ represents the conversation history, $c$ represents the candidate response, and $u_1, a_1, ..., u_i, a_i$ represents the conversation trajectory for the next i rounds. In this context, $u$ means a patient's statement and $a$ means a physician's statement. The variable $n$ acts as the upper limit for the length of the trajectory of interest. The discount factor $d \in (0, 1]$ indicates the level of importance assigned to the future impact.

Next, we will show the process for evaluating conversation states. To ensure clarity, we introduce the variable $h'$ to represent any conversation history that ends with a doctor's response. The corresponding status score for that conversation history can be determined using the following formula:
\begin{gather}
v_{h'}^r = \frac{1}{k} \sum_{k} \text{REM}(h', r) \\
v(h') = \sum_{r \in \mathcal{R}^g \cup \mathcal{R}^c} w_rv_{h'}^r
\end{gather} 
In the above equations, $r$ denotes the rule, and due to the somewhat random nature of scoring the REM, we make $k$ predictions and calculate the average as the final score. The weight of a rule, denoted $w_r$, is designed to indicate the order and level of goal accomplishment. For constrained rules, since the order is not considered, we assign a constant weight, denoted $\gamma$, which is set to a value close to 0 to emphasize the importance of prioritizing goal satisfaction over constraint satisfaction. For goal-oriented rules, we use the following formula to compute the weights: 
\begin{equation}
w_{r \in \mathcal{R}^g}=\prod_{r' \prec r} \mathcal{V}^{t_1}_{\alpha}(v_{h'}^{r'}) \prod_{r' \rightarrow r} \mathcal{V}^{t_2}_{\beta}(v_{h'}^{r'}) \label{eq:score}
\end{equation} 
In Equation \ref{eq:score}, $r' \prec r$ indicates that $r'$ is a predecessor rule to $r$, while $r' \rightarrow r$ indicates that $r'$ is a constraint rule for $r$. For example, in Figure~\ref{fig:flow}, Rule A is a predecessor rule to Rule B, Rule D is a constraint rule for Rules B and C, and Rules E and F are constraint rules for all goal-oriented rules. The function $\mathcal{V}^t_y(x), y \in [0, 1)$ represents a value function that equals 1 when $x \ge t$, and $y$ otherwise. We assign a value close to 0 to $\alpha$ to indicate that $r$ is less likely to be satisfied if its predecessor rule is not satisfied. Similarly, we assign a value close to 1 to $\beta$ to indicate that $r$ is still partially credible when its constraints are not satisfied.

\subsection{Human Preference Alignment}
In this subsection, we describe the process of training the model using preference data. The training process consists of two steps. Firstly, we fine-tune the base model using the dialogue dataset to enhance the model's medical conversation capabilities. In the second step, we utilize the DPO algorithm~\cite{dpo} to help the model learn from the preference data. The objective of this algorithm is to minimize the following expression: 
\begin{equation} 
\min_{\pi} - \mathbb{E}[\log \sigma(\lambda \log \frac{\pi(o_c \mid h)\pi_{\text{sft}}(o_r \mid h)}{\pi_{\text{sft}}(o_c \mid h)\pi(o_r \mid h)})] 
\end{equation} 
We label the fine-tuned model with the instructions as $\pi_\text{sft}$ and the model needing optimization with the same initial parameters as $\pi_\text{sft}$ as $\pi$. $o_c$ represents selected responses and $o_r$ represents rejected responses. The sigmoid activation function is denoted as $\sigma$, and $\lambda \in (0, 1)$ determines the difference between $\pi$ and $\pi_\text{sft}$, with a smaller $\lambda$ resulting in a larger difference.

\begin{table}[ht]
\centering
\scalebox{0.9}{
\begin{tabular}{@{}lll@{}}
\toprule
Statistic & Item & Value \\ \midrule
\multirow{2}{*}{Count} & Department Num & 5 \\
 & Case Num & 72 \\ \midrule
Avg Length & Patient Info & 493.3 \\ \midrule
\multirow{4}{*}{Avg Num} & QA pairs & 37.3 \\
 & Major Symptoms & 7.3 \\
 & Major Medical Test & 2.8 \\
 & Diseases & 1.7 \\ \bottomrule
\end{tabular}
}
\caption{
Statistics for the CSPT dataset. 
Patient Information describes the patient, while QA pairs represent doctor-patient questions and answers in dialog scripts, used to create simulated patients. Major Symptoms, Major Medical Test, and Diseases are also used to evaluate the model's dialog capability.
}
\label{table:sp_testing}
\end{table}

\section{Experiments}
\subsection{Standardized Patient Test}
In the realm of medicine, Standardized Patients (SP) imitate genuine patient symptoms and reactions following adequate training. Their portrayal of patient responses must be consistent and precise. When undergoing evaluation, standardized patients typically adopt a non-active communication approach, refraining from actively conveying information to the physician. This approach is employed to assess the physician's communication skills.
The development of LLMs has made it feasible to employ computer-simulated standardized patients. Some previous studies~\cite{huatuogpt-2023, wei-etal-2018-task} have aimed to evaluate the performance of models using similar approaches. However, these studies usually provide the modeled patients with limited symptom information, which often hinders the model's ability to accurately comprehend the patient's interaction with the doctor. In the medical field, this challenge is commonly addressed by instructing the standardized patient to memorize a detailed dialogue script, crafted by a professional, that realistically showcases the patient's responses to various inquiries. During the examination, the patient can then respond to the doctor in accordance with the script, ensuring the quality of the responses.

We created the Chinese Standardized Patient Testing~(CSPT). dataset on the basis of this idea. To simulate patients, we propose using the retrieval -augmented generation technique and inputting patient descriptions and dialogue scripts into a database. We used patient cases from the book "Objective Structured Clinical Examinations \& Standardized Patients"~\cite{sp_testing} for this purpose. The primary focus of our SP test is to gather key symptoms and medical tests, as well as accurately diagnose diseases. To assist in this, we have provided a reference list for each case. Table~\ref{table:sp_testing} showcases the dataset statistics, and the engineering implementation of the patient simulator can be found in Appendix~\ref{sec:sp_testing}.

\begin{table*}[ht]
\centering
\scalebox{0.90}{
\begin{tabular}{l|ccc|ccc|ccc|ccc}
\toprule
\multicolumn{1}{c|}{} & \multicolumn{3}{c|}{\textbf{Internal Medicine}} & \multicolumn{3}{c|}{\textbf{Surgery}} & \multicolumn{3}{c|}{\textbf{Other}} & \multicolumn{3}{c}{\textbf{ALL}} \\
\multicolumn{1}{c|}{\multirow{-2}{*}{\textbf{Model}}} & Sym. & Test & Dis. & Sym. & Test & Dis. & Sym. & Test & Dis. & Sym. & Test & Dis. \\ \hline \hline
Baichuan-Chat & 22.0 & 34.8 & 46.4 & 25.8 & 18.1 & 34.1 & 17.7 & {\color[HTML]{FD6864} \textbf{29.3}} & 22.4 & 21.7 & 27.4 & 33.8 \\
ChatGLM3 & 2.4 & 25.3 & 28.9 & 0.0 & 30.4 & 28.9 & 6.1 & 15.9 & 12.2 & 3.0 & 23.6 & 22.9 \\
Huatuo-II & 2.7 & 43.5 & 34.1 & 9.0 & 51.4 & 50.7 & 5.5 & 26.2 & 37.8 & 5.7 & 39.8 & 40.7 \\
DISC-MedLLM & {\color[HTML]{FD6864} \textbf{25.5}} & 30.2 & 50.0 & {\color[HTML]{FD6864} \textbf{30.6}} & 45.3 & 45.3 & 19.0 & 21.2 & 30.1 & {\color[HTML]{009901} \textbf{24.8}} & 34.9 & 41.3 \\ \hline \hline
SFT~(Qwen) & 13.4 & 32.2 & 39.9 & 16.2 & 40.9 & 35.5 & 17.7 & 18.6 & 26.9 & 15.9 & 30.1 & 33.8 \\
SFT~(Baichuan) & 17.2 & 30.9 & 40.1 & 14.6 & 38.9 & {\color[HTML]{009901} \textbf{57.2}} & 2.0 & 11.5 & 21.8 & 10.8 & 26.3 & 39.1 \\ \hline 
PLPF~(Qwen) & 19.8 & {\color[HTML]{FD6864} \textbf{47.8}} & {\color[HTML]{009901} \textbf{53.6}} & {\color[HTML]{009901} \textbf{29.5}} & {\color[HTML]{009901} \textbf{57.2}} & {\color[HTML]{009901} \textbf{57.2}} & {\color[HTML]{FD6864} \textbf{28.0}} & {\color[HTML]{009901} \textbf{29.2}} & {\color[HTML]{FD6864} \textbf{52.6}} & {\color[HTML]{FD6864} \textbf{25.9}} & {\color[HTML]{FD6864} \textbf{44.1}} & {\color[HTML]{009901} \textbf{54.4}} \\
PLPF~(Baichuan) & {\color[HTML]{009901} \textbf{24.8}} & {\color[HTML]{009901} \textbf{46.7}} & {\color[HTML]{FD6864} \textbf{64.5}} & 28.7 & {\color[HTML]{FD6864} \textbf{59.4}} & {\color[HTML]{FD6864} \textbf{66.7}} & {\color[HTML]{009901} \textbf{19.2}} & 20.0 & {\color[HTML]{009901} \textbf{41.0}} & 24.1 & {\color[HTML]{009901} \textbf{41.1}} & {\color[HTML]{FD6864} \textbf{56.7}} \\ \bottomrule
\end{tabular}
}
\caption{
The experiment results on the CSPT dataset.
The Symptom (Sym.) and Test metrics indicate the probability of the model identifying key symptoms and key medical tests, respectively, while the Diagnosis (Dis.) metric indicates the probability of the model making a correct diagnosis. 
We use red and green labels to denote the best and second-best results, respectively.
}
\label{table:main_result}
\end{table*}

\subsection{Test Settings}
The evaluation involves a simulated interaction between a model and a patient simulator. Two doctors assess the interaction using a predetermined checklist. The model's performance is measured as a percentage based on the successful completion of checklist items. The assessment procedure includes limiting the dialogue rounds to five and requiring the model to inquire about symptoms, provide a diagnosis, and propose a treatment plan within this time frame. It is worth mentioning that all models are evaluated under identical conditions, with a decoding temperature of 0, to ensure consistency in the assessment process.

\subsection{Baselines and implementation details}
We utilized different models as baselines, which can be classified into three groups: 1) Chat LLMs, such as ChatGLM3 (6B)~\cite{du2022glm, zeng2022glm} and Baichuan2-Chat (7B)~\cite{yang2023baichuan}; 2) Medical LLMs, including DISC-MedLLM~\cite{bao2023disc} and Huatuo-II~\cite{chen2023huatuogpt}; and 3) Instruction-tuned LLMs constructed on different backbones, specifically SFT~(Qwen~\cite{bai2023qwen}) and SFT~(Baichuan). The instruction data used is the same as that of DISC-MedLLM.  We implemented the PLPF model based on the SFT model. To ensure a fair comparison with DISC-MedLLM, we used an equal amount of data for preference learning. For additional information regarding baselines and the specific hyperparameters used for training the model, please consult Appendix \ref{sec:train} and \ref{sec:baseline}.

\begin{figure}[t]
    \centering
    \includegraphics[width=0.4\textwidth]{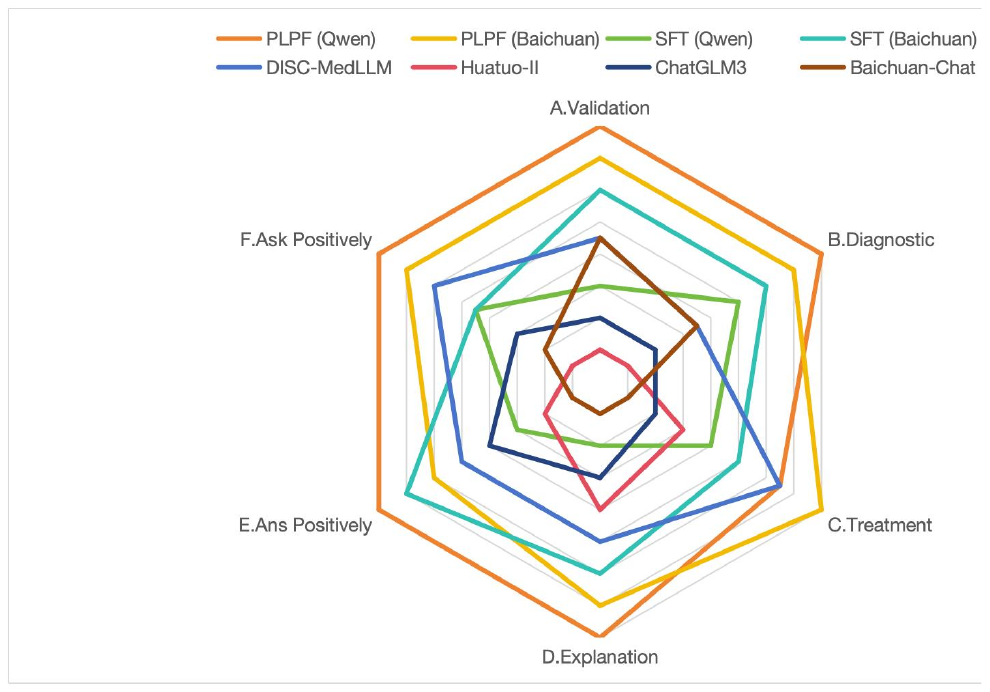}
    \caption{
    Ranking of how well each model follows the rules. Each axis of the radar graph corresponds to a rule in Fig.~\ref{fig:flow}, and we use the letters A-F to denote the mapping between rules and axes.
    }
    \label{fig:radar}
\end{figure}

\subsection{Experimental Results}
Table ~\ref{table:main_result} shows the overall results. To provide a comprehensive understanding of different models, we utilized REM~\footnote{Appendix~\ref{sec:rem} contains the performance evaluation of REM.} to assess the compliance level of these models with the rules depicted in Fig.~\ref{fig:flow}. We then generated a radar chart, as shown in Fig.~\ref{fig:radar}, which represents the ranking of compliance scores.

ChatGLM3 and Huatuo-II perform poorly in symptom collection, indicating a lack of active information request from the patient during communication. 
However, despite this limitation, Huatuo-II has the ability to recommend numerous medical tests to the patient during the dialogue. By analyzing the results of these tests, Huatuo-II still achieves a high rate of correct diagnosis. On the other hand, Baichuan2-Chat and ChatGLM3 have lower diagnostic accuracy because they often violate Rule E~(Ans Positively) when interacting with the user, rejecting the diagnosis by stating that they are AI models. SFT~(Qwen) and SFT~(Baichuan) score moderately on the indicators, placing them in the middle range in terms of adherence to the rules. Among these, SFT~(Qwen) violates Rule A~(Validation) and Rule E~(Ans Positively) more frequently, leading to lower correctness in disease diagnosis compared to SFT~(baichuan).

We will now examine DISC-MedLLM, which employs the same base model and fine-tuning dataset as our approach but differs in the preference data utilized. As shown in Figure~\ref{fig:radar}, DISC-MedLLM is more proactive in requesting information from patients and offering a wider range of treatment options compared to SFT~(Baichuan). However, the model exhibits less confidence in making diagnoses, as evidenced by its lower adherence to Rule B (Diagnose) and Rule E (Ans Positively). This limits the model's ability to effectively utilize its advantage of requesting more patient information, resulting in only a slightly higher correct diagnosis rate compared to SFT~(Baichuan). In contrast, the PLPF-optimized model shows a significant improvement of more than 10 points in the identification of symptoms, medical tests, and diseases compared to the SFT model, which emphasizes the effectiveness of our approach.
Please refer to Appendix~\ref{sec:case_study} for case studies.

\section{Analyze}
\subsection{Ablation Study}
In this subsection, we will further investigate how the scoring method (Eq.~\ref{eq:score}) and the trajectory prediction step (Sec.~\ref{sec:tp}) affect the model's ability to engage in multi-round conversations. To validate Eq.~\ref{eq:score}, we compare it with the weighting method by directly setting $w_r = 1$. Additionally, we vary the trajectory length to 1, 2, and 3 to assess the effectiveness of the trajectory prediction step. A trajectory length of 1 means that we only predict the immediate doctor responses.

We conducted studies using SFT (Baichuan) and presented the results in Table~\ref{table:ab_result}. Based on the results, we observed that our method performs better than the direct summation of all the rule scores. The direct summation approach leads to a model that lacks proficiency in inquiring about patients' symptoms, resulting in decreased diagnosis accuracy. This is because the rule encouraging symptom collection is only one of six rules, and directly adding up rule scores would diminish its impact. Additionally, increasing the trajectory prediction length helps the model understand the entire conversation flow. Specifically, extending the trajectory length from 2 to 3 resulted in a significant improvement in all aspects of the model's capabilities.

\begin{table}[t]
\centering
\setlength{\tabcolsep}{10pt}
\scalebox{0.8}{
\begin{tabular}{ll|ccc}
\toprule
\multicolumn{2}{c|}{\multirow{2}{*}{\textbf{Strategy}}} & \multicolumn{3}{c}{\textbf{Trajectory Length}} \\
\multicolumn{2}{l|}{} & \textit{k=1}   & \textit{k=2} & \textit{k=3} \\ \hline
\multirow{3}{*}{Avg} & Sym. & 5.1 & 9.5 & 19.7 \\
 & Test & 40.7 & 28.9 & 31.3 \\
 & Dis. & 43.7 & 36.3 & 42.6 \\ \hline
\multirow{3}{*}{Ours} & Sym. & 18.9 & 21.6 & {\color[HTML]{FD6864}\textbf{24.1}} \\
 & Test & 28.0 & 32.0 & {\color[HTML]{FD6864}\textbf{41.1}} \\
 & Dis. & 47.7 & 49.5 & {\color[HTML]{FD6864}\textbf{56.7}} \\ \bottomrule
\end{tabular}
}
\caption{Ablation test results.}
\label{table:ab_result}
\end{table}

\subsection{Performance on Public Datasets}
\begin{table}[t]
\centering
\scalebox{0.75}{
\begin{tabular}{l|ccc}
\toprule
Model & Meddg $\downarrow$ & Imcs $\downarrow$ & wMedQA $\downarrow$ \\ \midrule \midrule
Baichuan-Chat & {\color[HTML]{000000} 1.69} & {\color[HTML]{000000} 1.77} & 1.44 \\
ChatGLM3 & 1.66 & 1.73 & 1.23 \\
Huatuo-II & 1.60 & 1.73 & {\color[HTML]{FD6864} \textbf{1.11}} \\
DISC-MedLLM & 1.59 & 1.63 & 1.21 \\ \midrule \midrule
SFT~(Qwen) & 1.60 & 1.71 & 1.26 \\
SFT~(Baichuan) & 1.67 & 1.71 & 1.26 \\ \hline
PLPF~(Qwen) & {\color[HTML]{009901} \textbf{1.56}} & {\color[HTML]{FD6864} \textbf{1.59}} & {\color[HTML]{009901} \textbf{1.14}} \\
PLPF~(Baichuan) & {\color[HTML]{FD6864} \textbf{1.53}} & {\color[HTML]{009901} \textbf{1.60}} & 1.19 \\ \bottomrule
\end{tabular}
}
\caption{
Performance of the models on the Meddg, Imcs and WebMedQA dataset. 
We use red and green to highlight the best and second-best scores.
}
\label{table:pub}
\end{table}

\begin{figure}[t]
    \centering
    \includegraphics[width=0.45\textwidth]{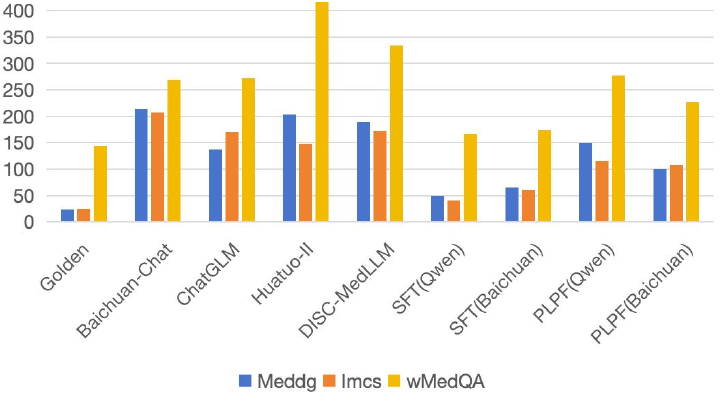}
    \caption{
    The average output length of LLMs over different datasets, where Goden represents the average length of the standard answer.
    }
    \label{fig:len}
\end{figure}
In this section, we report the results of all models on the public datasets. 
We selected the Meddg~\cite{liu2022meddg}, Imcs~\cite{imcs}, and WebMedQA~\cite{he2019applying} datasets to test the models' performance. 
The first two datasets are multi-turn dialogue datasets, while the last dataset consists of single-turn dialogue datasets.

During the evaluation of the model, we noticed a significant difference in the length of the content generated by the model compared to the standard answer, as shown in Fig.~\ref{fig:len}. As a result, traditional statistical measures like BLEU and ROUGE were not effective in evaluating the quality of the model's output. For more details, please refer to Appendix~\ref{sec:b@r}. We believe that the main focus for LLMs should be their ability to produce text that implies the standard answer, as this ensures the accuracy and dependability of the model's output. In accordance with this viewpoint, we have developed a new evaluation metric called GPT-Distance, which measures the extent to which the LLM output implies the standard answers. To be more specific, we utilized GPT-4 to determine whether the predictions imply the references, categorizing the level of implication as not implied, partially implied, or fully implied. The prompt used for this assessment is provided below:

\vspace{0.2cm}
\noindent \textit{Sentence 1: [predict]; Sentence 2: [reference] \\
Please decide if sentence 1 implies sentence 2? \\
A. Fully; B. Partially; C. Not.}
\vspace{0.2cm}

Subsequent to obtaining all predicted classifications, we calculate the GPT-Distance using the formula $(2 \times |\text{Not}| + |\text{Partially}|) / |\text{ALL}|$, where $|\text{Not}|$ and $|\text{Partially}|$ denote the number of samples categorized as not-implied and partially-implied, respectively. |\text{ALL}| indicates the number of test data.

We randomly selected 200 samples from each of the three datasets for testing, and the results are shown in Table~\ref{table:pub}.
Our findings demonstrate that the models trained by PLPF yielded the most favorable results for multi-round dialogs. 
Although Huatuo-II and DISC-MedLLM are able to generate longer responses, it is obvious that our generations have a higher coverage of physician responses, suggesting that PLPF allows the model to better understand the physician's diagnostic logic.
In the context of the single-round dialog task, Huatuo-II emerged as the top performer, with our model securing the second and third positions, respectively. 
This illustrates that while PLPF is primarily designed for multi-turn dialogue tasks, it also exhibits strong optimization for single-turn dialogue tasks.
\section{Related works}
\paragraph{Medical LLMs.}
Significant advancements have been made in medical dialogue models since the introduction of ChatGPT~\cite{chatgpt}. Recent research has primarily focused on creating large and high-quality instruction fine-tuning datasets for LLMs. Studies such as DoctorGLM~\cite{xiong2023doctorglm}, BenTsao~\cite{wang2023huatuo}, and ChatMed~\cite{zhu2023ChatMed} have proposed the utilization of powerful LLMs like ChatGPT to generate dialogue and question answering data at a low cost. However, ensuring the quality of data generated through this approach is challenging due to ChatGPT's tendency to generate incorrect information. To tackle this issue, Huatuo~\cite{huatuogpt-2023} suggested incorporating parts of real data into the generated data. Furthermore, to improve the readability of the real dialogue data, Huatuo refined it with ChatGPT, and this method has been widely adopted by subsequent research.
In addition to the dialogue data, several studies have aimed to generate various auxiliary task data. For instance, DISC-MedLLM\cite{bao2023disc} and ClinicalGPT\cite{wang2023clinicalgpt} have integrated knowledge graph-related data into the training data to enhance the model's ability to answer commonsense questions. ClinicalGPT has also attempted to improve the diagnostic capability of the model by including data from electronic medical records and medical examinations in training data. While there has been significant progress in fine-tuning medical LLM instructions, there is still limited research on the preferred learning stage.

\paragraph{Preference Learning.}
Preference alignment is a prominent focus in large model training research, as preferentially aligned models exhibit enhanced generalization ability in zero shot scenarios~\cite{kirk2023understanding}. 
Currently, the most renowned method for preference alignment is reinforcement learning from human feedback, which involves the utilization of four models for training. 
However, this approach has drawbacks, e.g., high engineering complexity and unstable training. 
In an effort to streamline the preference learning process, \citet{dpo} introduced a direct preference optimization algorithm that can bypass the need to train the reward model. 
Similarly, \citet{gulcehre2023reinforced} has proposed a self-reinforcement learning approach that uses the EM algorithm to eliminate the training of the critic model. 
Furthermore, RRHF\cite{yuan2023rrhf} suggests using learning rankings to replace reinforcement learning, thus strengthening learning stability. 
Furthermore, some initiatives, such as RLAIF\cite{lee2023rlaif}, aim to leverage AI to substitute manual preference data annotation, thus reducing annotation costs. 
Moreover, \citet{bai2022constitutional} proposes training constitutional evaluation models for self-reflection. 
\citet{sun2023salmon} has put forward the idea of using the Principle-following reward model as a replacement for the traditional reward model to achieve dynamic adaptation to human preferences.
Compared to these methods, our main contribution is to propose a preference learning approach for multiple rounds of dialogue.

\section{Conclusion}
In this work, we have introduced an innovative approach termed preference learning from process feedback (PLPF), which integrates the diagnostic logic of healthcare professionals into the LLM. 
PLPF encompasses rule modeling, preference data generation, and preference alignment to train the model to adhere to the diagnostic process. 
Our experimental findings, using standardized patient testing, reveal that PLPF enhances the diagnostic accuracy of the baseline model in medical conversations by 17.6\%. 
Furthermore, PLPF exhibits efficacy in both multi-round and single-round dialogue tasks, underscoring its potential for advancing medical dialogue generation.
\section*{Ethics and Limitations}
There are several limitations to our approach. Firstly, the defined processes are relatively simple, and more complex processes require additional testing. Secondly, the accuracy of the model is still not high enough for practical use in SP tests, as it sometimes generates hallucinatory results. Additionally, it is worth noting that there may be a geographical bias in the test results, as most of the cases used in our study came from the Wuhan region of China. Therefore, it is important to consider the ethical implications of this geographical bias when interpreting our findings.

\section{Acknowledgement}
This work is supported by the National Natural Science Foundation of China under Grant No. 62192731. We kindly appreciate all the researchers who provide valuable insights, discussions, and comments on this work.

\bibliography{anthology,custom}
\bibliographystyle{acl_natbib}

\appendix
\section{Rules Scoring Guidelines}
\label{appendix: scoring}
\begin{figure*}
    \centering
    \includegraphics[width=0.95\textwidth]{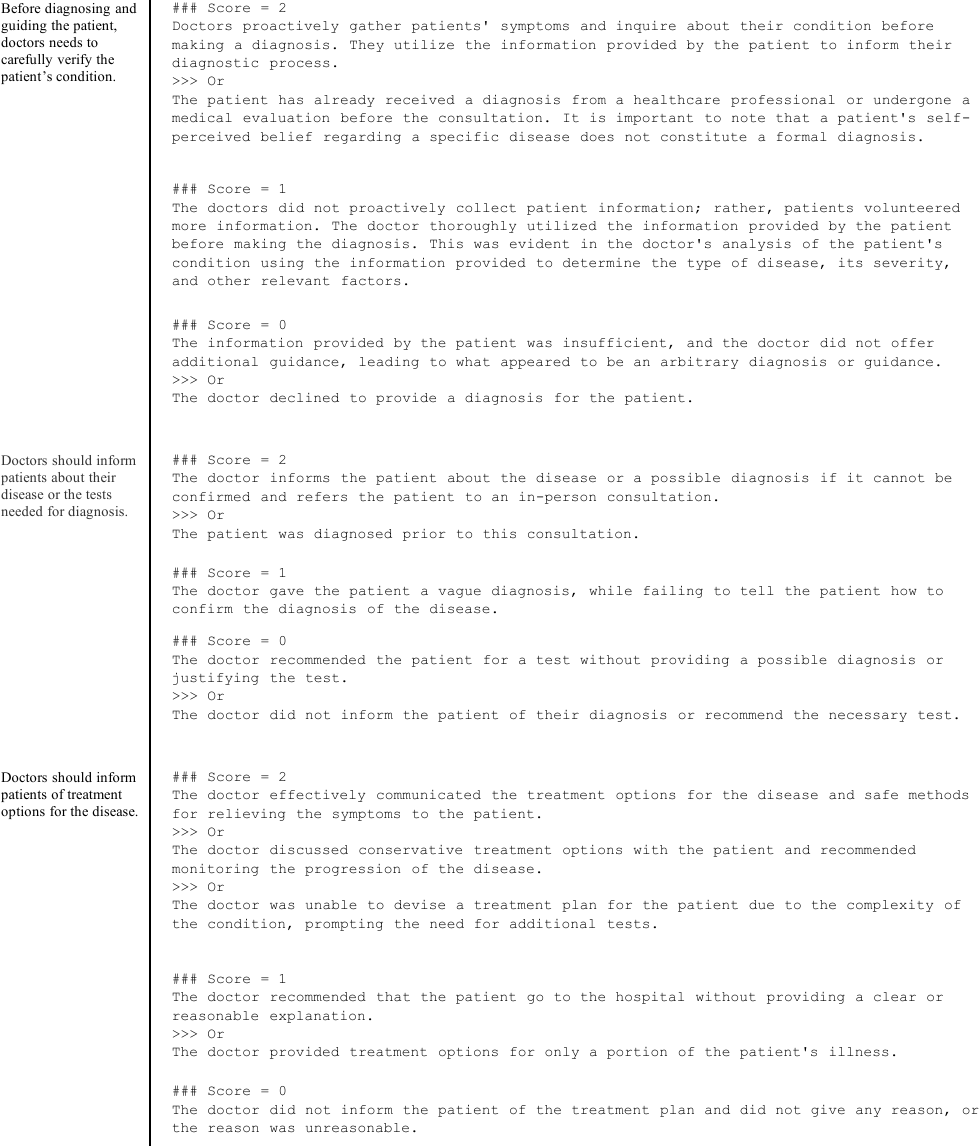}
    \caption{Goal-oriented rules evaluation criteria.}
    \label{fig:rules}
\end{figure*}

In this section, we provide the annotation guidelines for annotators' reference. We divide the rules into two categories: goal-oriented rules and constraint-oriented rules. For goal-oriented rules, we have defined strict scoring criteria, which are illustrated in Fig.~\ref{fig:rules}. However, assessing the level of compliance for constraint-oriented rules is difficult, so we allow annotators to score freely. In the end, we consider the consensus among the annotators as the final score.

To reduce the annotation workload, we utilized ChatGPT to assist in the annotation process. Specifically, we manually created specific scenarios for each score of each rule and employed the In-context Learning technique to allow ChatGPT to pre-label the data. The annotators' task was to review the annotations generated by ChatGPT and make necessary corrections. Our findings indicate that this correction-based approach significantly improves the internal consistency of the annotators.

\section{Standardized Patient Testing}
\label{sec:sp_testing}
\subsection{Data}
During the SP tests, we used three types of data: patient information, dialog scripts, and checklists. Patient information and dialog scripts were employed to create simulated patients, while checklists were used to assess the history of dialogs generated by the model following interactions with simulated patients. Our dataset included information from five departments and the number of cases in each department was as follows: 23 cases in internal medicine, 23 cases in surgery, 8 cases in gynecology, 10 cases in pediatrics, and 8 cases in psychiatry. The following section provides a detailed explanation of these three types of data.

\begin{figure*}
    \centering
    \includegraphics[width=0.95\textwidth]{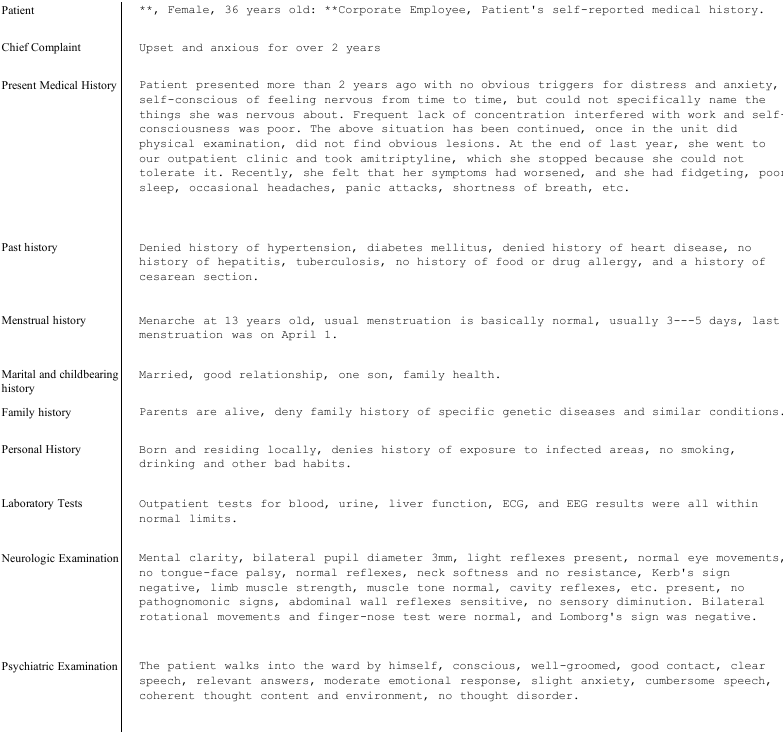}
    \caption{Example of Patient Information.}
    \label{fig:p_info}
\end{figure*}

\paragraph{Patient Information.}
The patient data consists of a wide range of information, such as patient symptoms and treatments, among other things. An example of patient information is shown in Fig.~\ref{fig:p_info}. Patient data includes a significant amount of laboratory test results, which can be used to assess the analytical capabilities of the model.

\begin{figure*}
    \centering
    \includegraphics[width=0.95\textwidth]{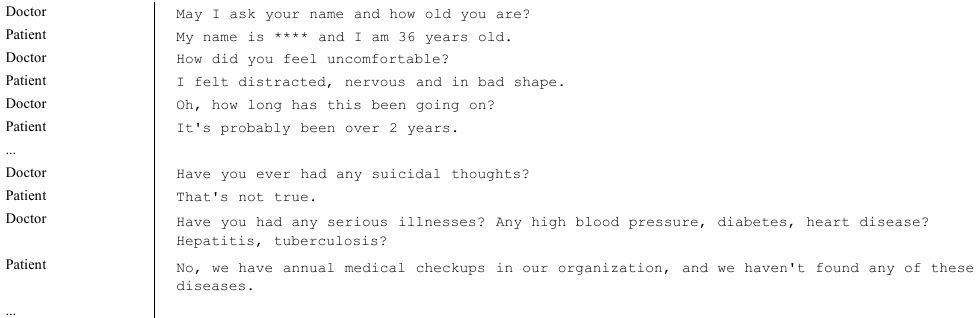}
    \caption{Example of Dialogue Script.}
    \label{fig:script}
\end{figure*}

\paragraph{Dialogue Script.}
Although the patient information is detailed, it does not capture patient mood, speech style, and life experience. To make the patient simulation more realistic, a dialogue script is provided (see Fig.~\ref{fig:script}). The script includes numerous exchanges between the doctor and the patient, involving both inquiries and responses. In addition to discussing important symptoms and tests, the script also includes inquiries and responses about less significant symptoms. These less significant symptoms act as distractors for the LLM test, thus improving the reliability of our test.

\begin{figure*}
    \centering
    \includegraphics[width=0.95\textwidth]{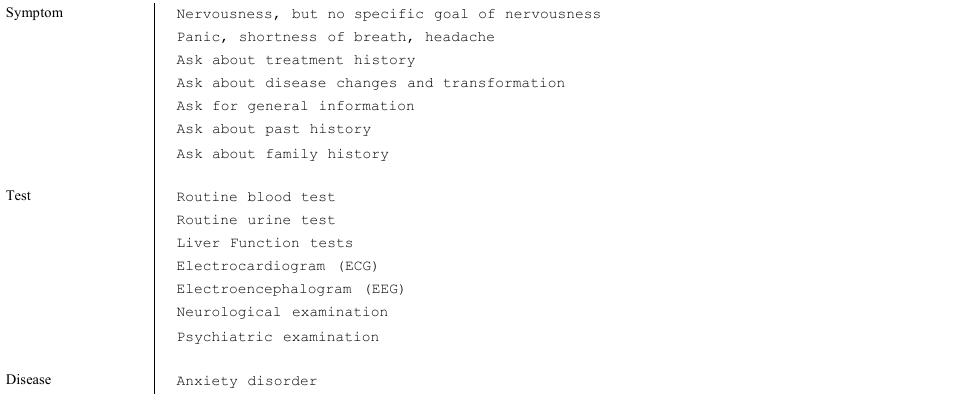}
    \caption{Example of Checklist.}
    \label{fig:checklist}
\end{figure*}

\paragraph{Checklist.}
A checklist is used to evaluate the model, which consists of three parts: key symptoms, key tests, and diseases, as shown in Fig.~\ref{fig:checklist}. Essentially, a better understanding of the key symptoms and key tests will improve the model's ability to provide a precise diagnosis. It should be noted that the evaluation of treatment history, family history, and other factors has been included in the symptom section.

\begin{figure}[th]
    \centering
    \includegraphics[width=0.48\textwidth]{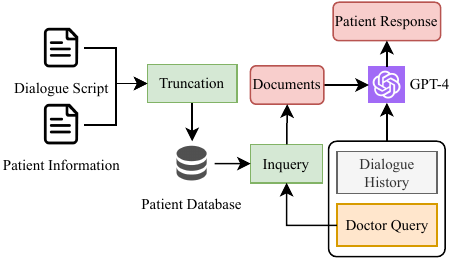}
    \caption{Patient Simulator Architecture.}
    \label{fig:sp}
\end{figure}

\subsection{Patient Simulator}
In carry out the testing process, we created a patient simulator to interact with the model being evaluated, as shown in Fig.~\ref{fig:sp}. The simulator was developed using the retrieval-augmented generation technique. Before conducting the tests, we created a separate database for each patient. Initially, we divided patient information and conversation scripts into documents with a maximum length of 128 tokens. Then, we encoded these documents using OpenAI's "text-embedding-ada-002" model to obtain a vector representation for indexing purposes. During the testing phase, we encoded the last two rounds of conversation history using the same "text-embedding-ada-002" model and retrieved the four most similar documents from the patient database to assist in generating responses. The model used to generate the responses was "gpt-4-0613", and the specific prompt used is shown below.

\vspace{0.2cm}
\noindent \textit{Please play the role of a patient who interacts with a doctor. You need to fulfill the following requirements: \\
1. If the doctor asks a question, answer it based on the contents of the knowledge base and the history of the conversation, with a response of no more than two sentences. \\
2. If your doctor recommends a test, inform him of the results. If you have not undergone the test, simply state that you are unaware of the results. \\
3. Do not expose any information about yourself to the doctor unless the doctor takes the initiative to ask a question, please be passive and accept the doctor's guidance. \\
4. If the doctor does not ask questions, ask the doctor what disease you have and how it should be treated. \\
5. When you feel that the conversation should end, please output: (End of Conversation). \\ \\
Knowledge Base: [documents] \\
Conversation history: [history] \\
Doctor: [question] \\
Your response: }
\vspace{0.2cm}

To avoid excessive interaction, we set a maximum of 5 rounds for communication between the LLM and the simulator during testing. This choice was made after noticing that medical LLMs usually need only 2-3 rounds of conversation to arrive at a patient's diagnosis.

\subsection{Evaluation}
We employ a manual evaluation method to assess each model using a checklist. To ensure the reliability of the assessment, two physicians will independently assign scores to each conversation history. The final score will be determined by calculating the average of their scores. The scoring formula we have adopted is as follows. 
\begin{equation}
s^j = \frac{1}{n} \sum_{i=1}^n \frac{c^j_i}{m^j_i} \quad j \in \{\text{Sym.}, \text{Test}, \text{Dis.}\}
\end{equation} 
Here, $n$ represents the number of standardized patients, and $j$ represents the specific assessment category, namely symptoms, medical tests, and diseases. $c^j_i$ represents the number of items that the model "passes" in the j-th assessment category for the i-th patient, while $m_i^j$ represents the total number of items in the j-th assessment category. 
A "pass" in this context means that: 1) The model actively requests a symptom or medical test result from the patient, and it is included in the checklist.
2) The model predicts a disease and is included in the checklist. It is important to note that if the model provides four or more candidate diseases simultaneously, we consider it a failure to diagnose the disease.

\section{Implementation of Our Model}
\label{sec:train}
All our models were trained using 4 A100-40G. We used the Lora~\cite{lora} technique in the training process, setting lora $\alpha$ and lora $r$ to 16 and 64, respectively, and the learning rate to 1e-4. 
For the Qwen model, we trained the modules "c\_attn" and "c\_proj", as well as "w1" and "w2". 
For the Baichuan model, the modules "W\_pack" and "o\_proj" were trained. 
The batch size used for training was 2, and the gradient accumulation steps were set to 16. 
Our REM was trained by Baichuan2-Chat (7B) using 1,800 samples, and the number of training rounds was 2. 
When calculating the score, we set the values of $\alpha$, $\beta$, $\gamma$, $d$, $t_1$, $t_2$ and $k$ to 0.1, 0.8, 0.1, 0.65, 1.0, 1.0 and 5, respectively. 

When applying REM to label preference data, response pairs are classified as win, tie, or loss, with a tie indicating that the difference between the scores of two responses is less than 1. 
After discarding all pairs labeled as tie, we select the top 2k samples with the largest difference to train our target model. 
Among preference data, the win-to-loss ratio between the trajectories obtained by data sampling and the tragjectory prediction is 1.2:1.

\section{Baselines}
\label{sec:baseline}
\paragraph{Chat.} 
1) ChatGLM3 (6B)~\cite{du2022glm, zeng2022glm}: This model is considered the most advanced Chinese LLM with a size within 10B and has shown performance comparable to GPT-4 in the OpenCompass~\footnote{https://opencompass.org.cn/leaderboard-llm} Chinese benchmark. 
2) Baichuan2-Chat (7B)~\cite{yang2023baichuan}: This model is based on Baichuan2-Base (7B), which is the preferred base model for Chinese medical LLMs in recent studies. Baichuan2-Chat (7B) performs similarly to ChatGPT on the OpenCompass Chinese benchmark.
\paragraph{Medical.} 
1) DISC-MedLLM~\cite{bao2023disc}: This model is based on Baichuan2-Base (13B) and was fine-tuned using 470k medical instruction data~\footnote{https://huggingface.co/datasets/Flmc/DISC-Med-SFT}. Additionally, this model utilizes 2k data for RLHF. 
2) Huatuo-II~\cite{chen2023huatuogpt}: This model is based on Baichuan2-Base (7B) and was fine-tuned using 5,252k pretraining instruction data and 142k medical Q\&A data. 
\paragraph{SFT.} To create two baselines, SFT~(Qwen) and SFT~(Baichuan), we used the same instruction fine-tuning data as DISC-MedLLM to fine-tune Qwen-Base (7B)~\cite{bai2023qwen} and Baichuan2-Base (7B), respectively.

\section{Performance of REM}
\label{sec:rem}
\begin{table}
\centering
\begin{tabular}{@{}lcc@{}}
\toprule
Model   & Exact Match & Fuzzy Match \\ \midrule
ChatGPT & 56.2        & 79.5        \\
Ours    & 62.1        & 88.3        \\ \bottomrule
\end{tabular}
\caption{Performance of different REMs on the testset.}
\label{table:rem}
\end{table}

In this section, we showcase the performance of REM on the test set using two different configurations. The initial configuration involves exact matching, where we determine the percentage of samples that REM accurately scores. The second configuration involves fuzzy matching, where we assess the likelihood that REM misclassifies a sample with a score of 2 as 0. and vice versa. A high score on this metric indicates that REM effectively distinguishes between good and bad responses. We conducted a comparison between REM, ChatGPT, and 5 instances manually created for In-context Learning to enhance ChatGPT's accuracy. The results of the experiment are presented in Table~\ref{table:rem}. Fine-tuned REM exhibits superior performance compared to ChatGPT. However, given the limited performance gap, it is expected that as the overall performance of the generalized LLM improves in the future, the entire PLPF process will become automated, with human intervention only required to design the flow chart and write the rules.

\section{Case Study}
\label{sec:case_study}
In this section, we will analyze the response preferences of each model during multi-round conversations, with the assistance of several examples. 
We have chosen multiple models for our analysis, all of which communicate with the same standardized patient suffering from acute appendicitis, using Baichuan-Base as the base model.

\begin{figure*}
    \centering
    \includegraphics[width=0.95\textwidth]{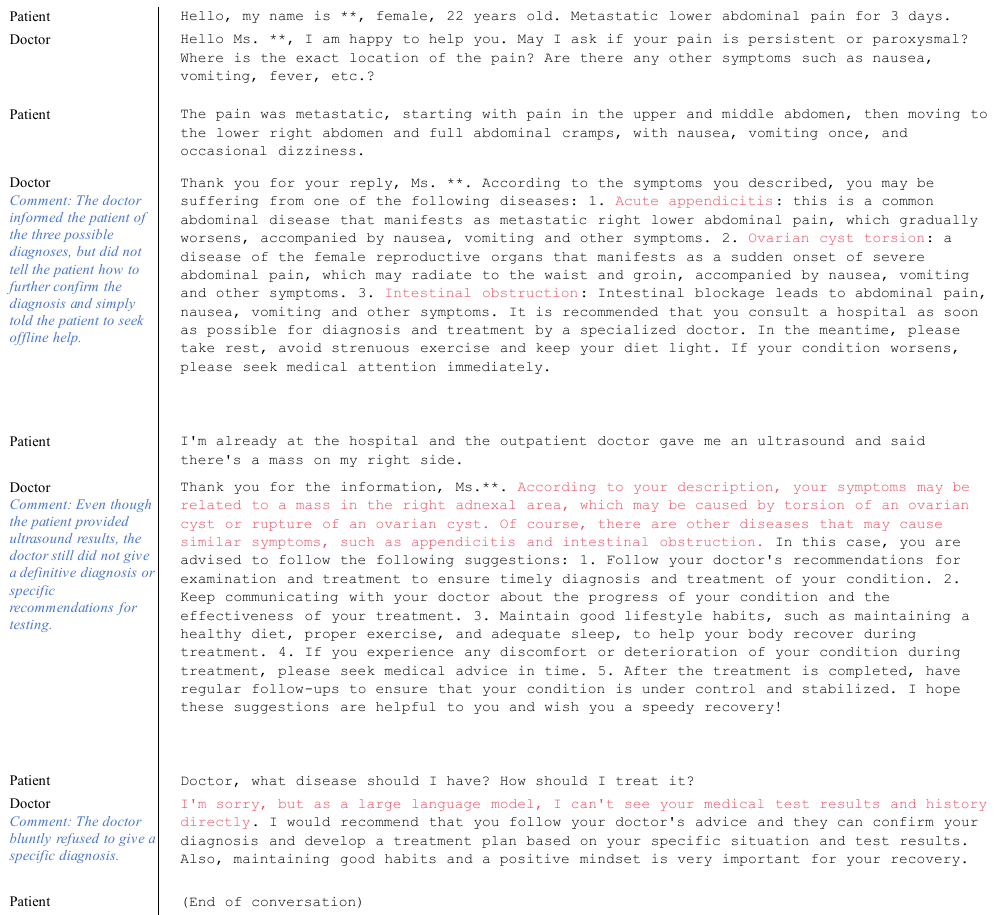}
    \caption{Example of Baichuan-Chat.}
    \label{fig:baichuan}
\end{figure*}

\paragraph{Baichuan-Chat.}
The conversation history of Baichuan-Chat is shown in Fig.~\ref{fig:baichuan}. Our analysis indicates that the model successfully generates an extensive range of potential patient diagnoses and links them to detailed explanations. However, the model lacks in providing guidance to the patient on how to confirm the diagnosis, and it also tends to avoid answering certain patient questions. As a result, these limitations reduce the diagnostic effectiveness of Baichuan-Chat.

\begin{figure*}
    \centering
    \includegraphics[width=0.95\textwidth]{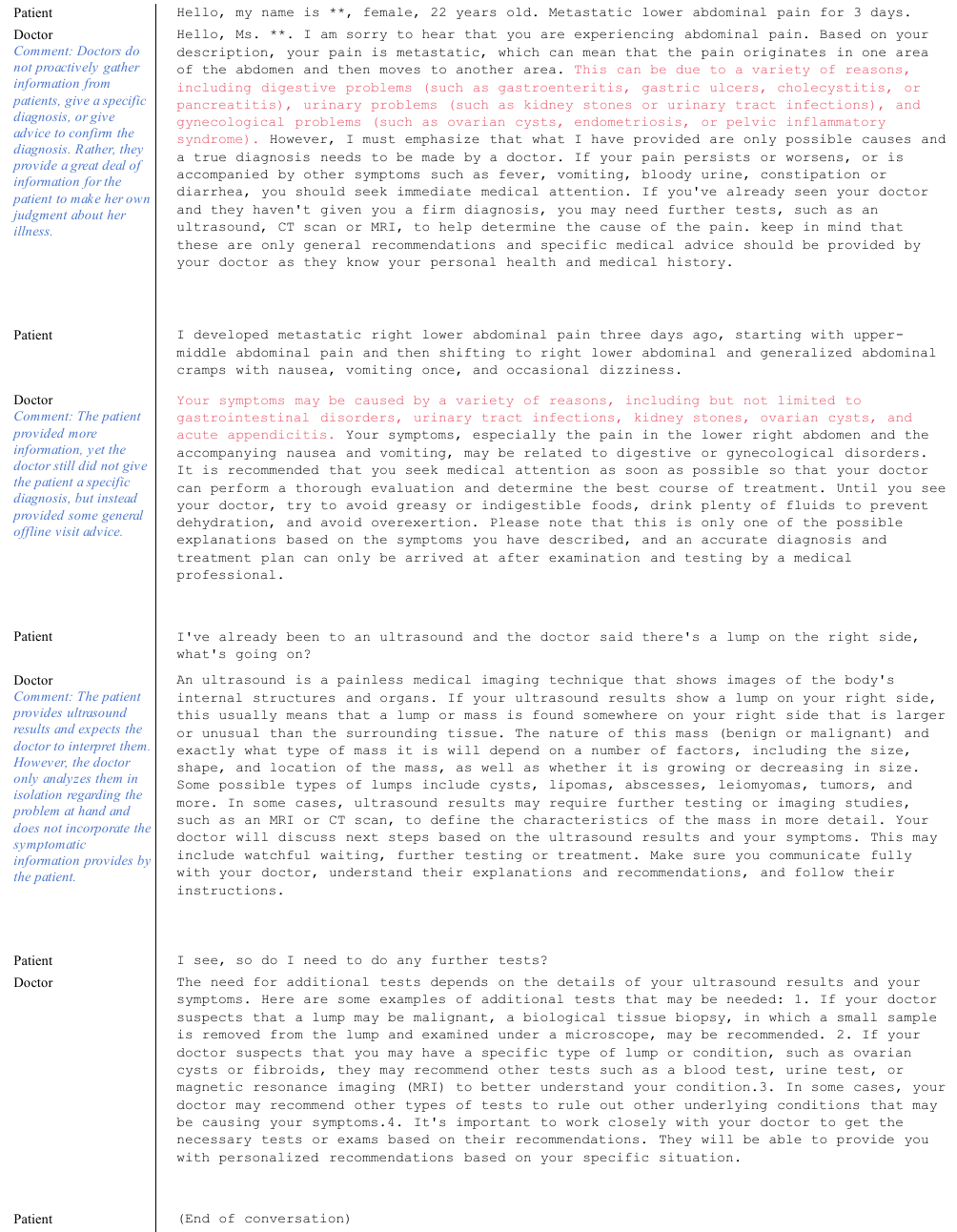}
    \caption{Example of Huatuo-II.}
    \label{fig:huatuo}
\end{figure*}

\paragraph{Huatuo-II.}
Fig.~\ref{fig:huatuo} presents the conversation history of Huatuo-II, which is characterized by its utilization of single-round conversations to achieve multi-round conversational objectives. A notable limitation of Huatuo-II is its inability to aid patients in interpreting medical test results by incorporating symptom information from previous dialogues. Moreover, Huatuo-II adopts a passive interaction style, overwhelming patients with an excessive amount of information that may impede their ability to extract valuable insights from the system's responses.

\begin{figure*}
    \centering
    \includegraphics[width=0.95\textwidth]{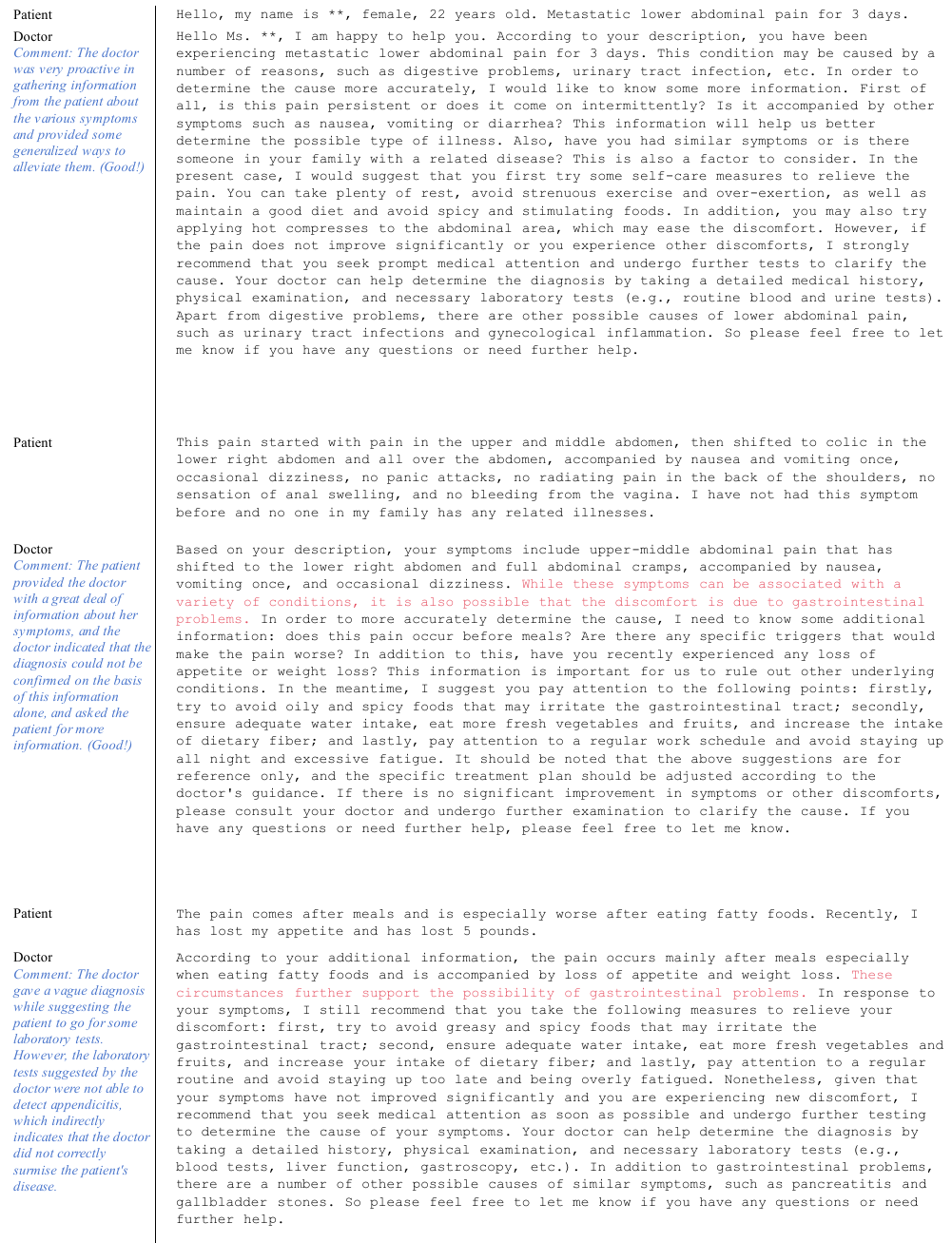}
    \caption{Example of DISC-MedLLM, Part-I.}
    \label{fig:disc-I}
\end{figure*}

\begin{figure*}
    \centering
    \includegraphics[width=0.9\textwidth, page=2]{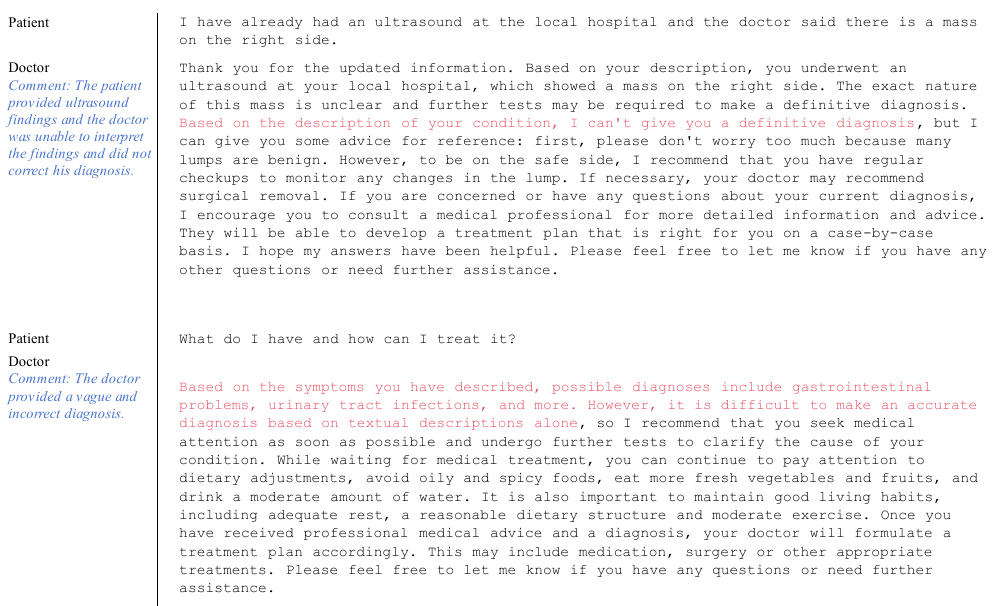}
    \caption{Example of DISC-MedLLM, Part-II.}
    \label{fig:disc-II}
\end{figure*}

\paragraph{DISC-MedLLM.}
The conversation history of DISC-MedLLM is presented in Fig.~\ref{fig:disc-I} and Fig.~\ref{fig:disc-II}. Our analysis suggests that DISC-MedLLM effectively extracts information from patients regarding their symptoms. However, the model relies on a fixed response template, where it restates the patient's statement, provides its own perspective, and concludes with recommendations, with a significant portion of the response dedicated to offering suggestions. As a result, DISC-MedLLM's responses tend to be longer compared to other models. One major drawback of DISC-MedLLM is the lack of specificity in its points. For instance, when diagnosing a patient with appendicitis, the model simply suggests that the patient may be experiencing a gastrointestinal issue. Additionally, while DISC-MedLLM provides numerous therapeutic recommendations, they are general in nature and do not offer comprehensive guidance.

\begin{figure*}
    \centering
    \includegraphics[width=0.9\textwidth]{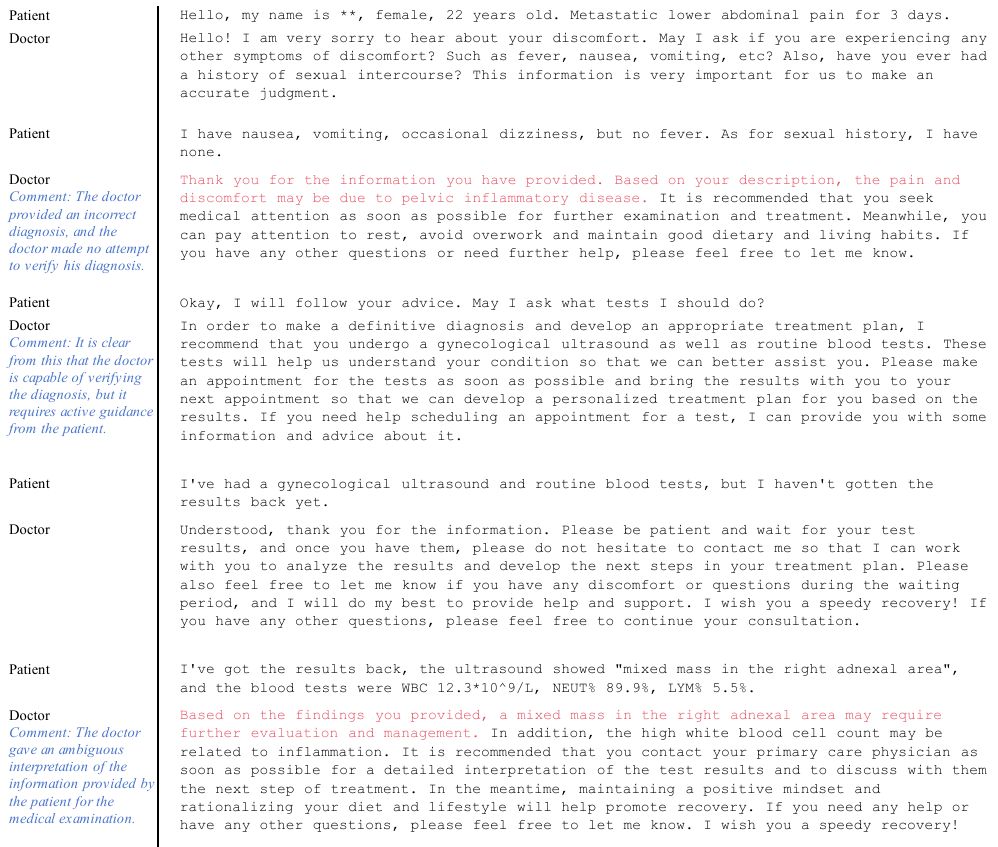}
    \caption{Example of SFT~(Baichuan).}
    \label{fig:sft}
\end{figure*}

\paragraph{SFT~(Baichuan)}
The conversation history of SFT~(Baichuan) is shown in Fig.~\ref{fig:sft}. Our observations indicate that SFT~(Baichuan) and Baichuan-Chat both fail to effectively provide patients with information on how to confirm their diagnosis. Furthermore, SFT~(Baichuan) analyzes the test results submitted by patients in a similar manner to DISC-MedLLM, as it advises patients that further evaluation of the test results is necessary, but both lacking detailed analysis of the test results. From this we can infer that DISC-MedLLM primarily improves the model's ability to inquire about symptoms and offer treatment recommendations.

\begin{figure*}
    \centering
    \includegraphics[width=0.95\textwidth]{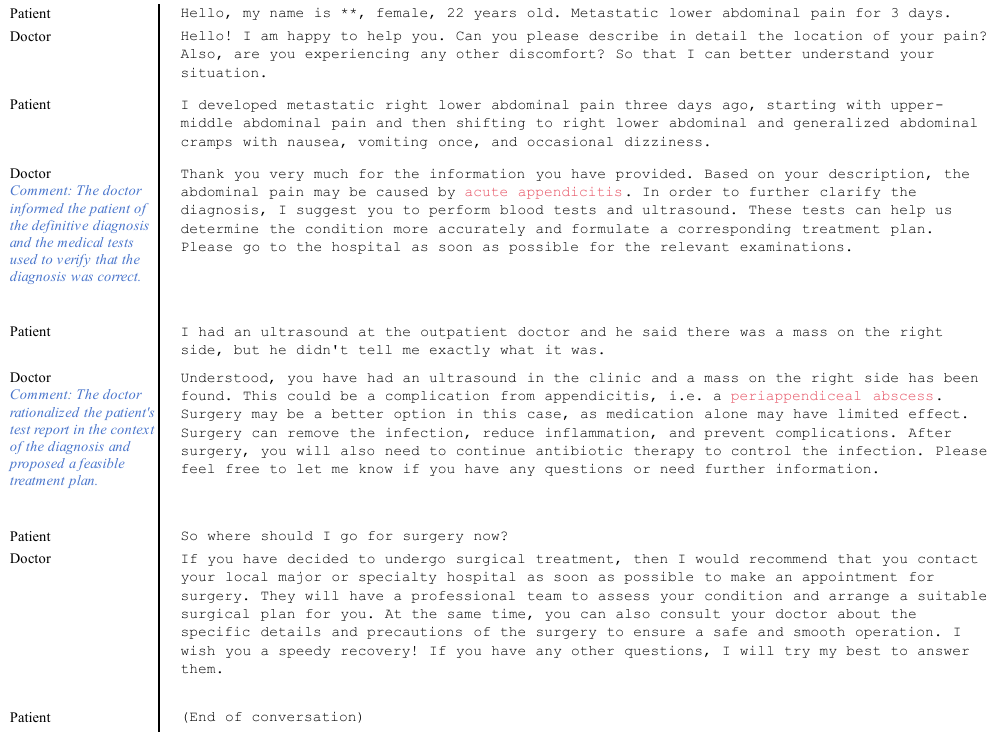}
    \caption{Example of PLPF~(Baichuan).}
    \label{fig:plpf}
\end{figure*}

\paragraph{PLPF~(Baichuan)}
The conversation history of PLPF~(Baichuan) is shown in Fig.~\ref{fig:plpf}. The PLPF model strictly adheres to a process that involves asking for symptoms, proposing a diagnosis, verifying the diagnosis, and suggesting a treatment recommendation. In comparison to the SFT model, the PLPF model is more focused on symptom inquiry. For example, our model specifically asks about the location of pain when it identifies the keyword "Metastatic ... pain," which is important for determining the possibility of appendicitis in the patient. In terms of validating the diagnosis, our model suggests more precise tests such as blood tests and ultrasound, while DISC-MedLLM suggests more general tests like gastroscopy and liver function tests. Our model effectively utilizes the findings of test results to further refine the patient's diagnosis, specifically identifying the possibility of septic appendicitis. On the contrary, the other models do not effectively utilize this information. Lastly, when it comes to offering treatment options, our model proposes a surgical treatment plan, whereas the other LLMs only provide a generic treatment plan.

\begin{table*}[htbp]
\centering
\small
\scalebox{0.9}{
\begin{tabular}{l|ccc|ccc|ccc}
\toprule
 & \multicolumn{3}{c|}{Meddg(Avg Len = 23.7)} & \multicolumn{3}{c|}{IMCS(Avg Len = 25.2)} & \multicolumn{3}{l}{WebMedQA(Avg Len = 144.3)} \\
\multirow{-2}{*}{Model} & B@4 & R@L & Len & B@4 & R@L & Len & B@4 & R@L & Len \\ \midrule \midrule
Baichuan-Chat & {\color[HTML]{6200C9} \textbf{0.5}} & {\color[HTML]{6200C9} \textbf{4.4/34.2/6.8}} & {\color[HTML]{FE0000} \textbf{214.2}} & {\color[HTML]{6200C9} \textbf{0.5}} & {\color[HTML]{6200C9} \textbf{4.3/27.9/6.3}} & 269.4 & {\color[HTML]{6200C9} \textbf{2.3}} & {\color[HTML]{6200C9} \textbf{9.7/18.9/10.5}} & 269.4 \\
ChatGLM3 & {\color[HTML]{FE0000} \textbf{1.8}} & {\color[HTML]{FE0000} \textbf{14.4/26.1/14.7}} & 138.0 & {\color[HTML]{FE0000} \textbf{1.6}} & {\color[HTML]{FE0000} \textbf{13.2/20.5/12.2}} & {\color[HTML]{333333} 272.2} & 3.4 & 11.2/22.7/12.8 & 272.2 \\
Huatuo-II & {\color[HTML]{6200C9} \textbf{0.6}} & {\color[HTML]{6200C9} \textbf{8.6/33.2/10.1}} & {\color[HTML]{FE0000} \textbf{188.7}} & {\color[HTML]{6200C9} \textbf{0.7}} & {\color[HTML]{6200C9} \textbf{7.4/37.3/9.2}} & {\color[HTML]{FE0000} \textbf{333.8}} & {\color[HTML]{6200C9} \textbf{2.9}} & {\color[HTML]{6200C9} \textbf{8.1/28.1/11.4}} & {\color[HTML]{FE0000} \textbf{425.1}} \\
DISC-MedLLM & {\color[HTML]{6200C9} \textbf{0.7}} & {\color[HTML]{6200C9} \textbf{6.5/28.9/7.6}} & {\color[HTML]{FE0000} \textbf{203.2}} & {\color[HTML]{6200C9} \textbf{0.7}} & {\color[HTML]{6200C9} \textbf{9.5/25.0/9.0}} & {\color[HTML]{FE0000} \textbf{425.1}} & {\color[HTML]{6200C9} \textbf{2.8}} & {\color[HTML]{6200C9} \textbf{8.8/24.7/11.7}} & {\color[HTML]{FE0000} \textbf{333.8}} \\ \midrule \midrule
SFT (Qwen) & {\color[HTML]{FE0000} \textbf{2.0}} & {\color[HTML]{FE0000} \textbf{19.5/26.4/17.4}} & {\color[HTML]{6200C9} \textbf{50.0}} & {\color[HTML]{FE0000} \textbf{2.1}} & {\color[HTML]{FE0000} \textbf{17.8/18.5/13.0}} & {\color[HTML]{6200C9} \textbf{167.4}} & {\color[HTML]{FE0000} \textbf{4.3}} & {\color[HTML]{FE0000} \textbf{14.3/18.5/13.8}} & {\color[HTML]{6200C9} \textbf{167.4}} \\
SFT (Baichuan) & {\color[HTML]{FE0000} \textbf{1.5}} & {\color[HTML]{FE0000} \textbf{14.9/27.0/14.4}} & {\color[HTML]{6200C9} \textbf{64.4}} & {\color[HTML]{FE0000} \textbf{1.3}} & {\color[HTML]{FE0000} \textbf{14.3/19.5/11.3}} & {\color[HTML]{6200C9} \textbf{173.5}} & {\color[HTML]{FE0000} \textbf{4.0}} & {\color[HTML]{FE0000} \textbf{12.9/19.6/13.6}} & {\color[HTML]{6200C9} \textbf{173.5}} \\ \midrule
PLPF (Qwen) & 1.0 & 7.5/36.9/10.3 & 149.2 & 1.2 & 8.9/28.0/10.0 & {\color[HTML]{FE0000} \textbf{277.8}} & 3.5 & 10.4/23.8/12.8 & {\color[HTML]{FE0000} \textbf{277.8}} \\
PLPF (Baichuan) & 1.3 & 11.3/34.8/13.1 & {\color[HTML]{6200C9} \textbf{99.8}} & 1.1 & 10.4/26.7/10.3 & {\color[HTML]{6200C9} \textbf{227.3}} & {\color[HTML]{FE0000} \textbf{4.0}} & {\color[HTML]{FE0000} \textbf{12.5/21.5/13.8}} & {\color[HTML]{6200C9} \textbf{199.6}} \\ \bottomrule
\end{tabular}
}
\caption{BLUE and ROUGE scores of LLMs on each dataset.}
\label{table:blue_rouge}
\end{table*}

\section{Evaluation Based on BLUE and ROUGE}
\label{sec:b@r}
In this section, we provide the BLUE-4 and Rouge-L scores obtained by the model on the Meddg, IMCS, and webMedQA datasets. We also discuss the limitations associated with these scores. The scores are presented in Table~\ref{table:blue_rouge}, with the three highest scores highlighted in red and the three lowest scores in purple for each dataset. Moreover, we include the length of the responses generated by each LLM. Unlike conventional reporting methods that typically only present the F1-score, we present the precision, recall, and F1-score together for the ROUGE score, separated by the "/" sign.

A strong negative correlation was observed between the precision metrics scores (such as BLUE and Rouge-Precision) and the length of LLM responses. Specifically, Rouge-Precision has a significant impact on ROUGE-F1. When traditional metrics are used for evaluation, models with shorter response lengths tend to receive higher scores. One possible explanation for this finding is that physicians' responses in real datasets are usually more concise, while the output of medical LLMs often includes additional details that are not present in the reference responses. As a result, the BLUE and ROUGE-F1 scores are lower. It is clear that the evaluation of a response cannot solely rely on its length, indicating that BLUE and ROUGE are not reliable measures for assessing the performance of LLM responses.

\end{document}